\documentclass[]{fairmeta}
\usepackage{microtype}
\usepackage{graphicx}
\usepackage{booktabs} 
\usepackage{hyperref}

\usepackage{times}
\usepackage{latexsym}
\usepackage{algorithm,algorithmic}
\usepackage[T1]{fontenc}
\usepackage[utf8]{inputenc}
\usepackage{amsmath}
\usepackage{amssymb}
\usepackage{mathtools}
\usepackage{amsthm}
\usepackage{bm}
\usepackage{multirow}
\usepackage{float}
\usepackage{adjustbox}
\usepackage{makecell}
\usepackage{setspace}
\usepackage{xcolor, colortbl}
\definecolor{Gray}{gray}{0.9}
\usepackage{inconsolata}
\usepackage{url}
\usepackage{amsfonts,amssymb}
\usepackage{mathrsfs}
\usepackage{newfloat}
\usepackage{listings}
\usepackage{enumitem}
\usepackage{tabularx}
\usepackage{array}
\usepackage[justification=centering]{caption}
\usepackage{nicefrac} 
\usepackage{comment}
\usepackage{subcaption}
\usepackage{tablefootnote}
\usepackage{textcomp} 
\usepackage{scalerel}
\usepackage{tikz}
\usepackage{wrapfig}
\usepackage{longtable}

\theoremstyle{plain}

\theoremstyle{definition}

\theoremstyle{remark}

\usepackage[textsize=tiny]{todonotes}
\definecolor{MAEblue}{RGB}{47 112 182}
\definecolor{mydarkgreen}{RGB}{0, 139, 69}

\title{Diversity-driven Data Selection for Language Model Tuning through Sparse Autoencoder}

\author[]{Xianjun Yang}
\author[]{Shaoliang Nie}
\author[]{Lijuan Liu}
\author[]{Suchin Gururangan}
\author[]{Ujjwal Karn}
\author[]{Rui Hou}
\author[]{Madian Khabsa}
\author[]{Yuning Mao}

\affiliation[]{Meta GenAI}

\abstract{
Instruction tuning data are often quantity-saturated due to the large volume of data collection and fast model iteration, leaving data selection important but underexplored. Existing quality-driven data selection methods, such as LIMA (NeurIPS 2023 \citep{zhou2024lima}) and AlpaGasus (ICLR 2024 \citep{chenalpagasus}) generally ignore the equal importance of data diversity and complexity. In this work, we aim to design a diversity-aware data selection strategy and creatively propose using sparse autoencoders (SAEs) to tackle the challenge of data diversity measure. 
In addition, SAEs can also provide more interpretability of model behavior and explain, e.g., the surprising effectiveness of selecting the longest response (ICML 2024 \citep{zhaolong}). Using effective data selection, we experimentally prove that models trained on our selected data can outperform other methods in terms of model capabilities, reduce training cost, and potentially gain more control over model behaviors.
We prove that SAEs can serve as a good alternative to diversity measure and design our method to be scalable for potential industrial large-scale pruning, and we will also release our trained SAEs for use by the broader community.}

\date{\today}
\correspondence{\email{xianjunyang@meta.com}, \email{yuningm@meta.com}}

\newcommand{\one}{SAE-GreedSelect }
\newcommand{\two}{SAE-SimScale }

\begin{document}

\maketitle

\section{Introduction}
Scaling large language models (LLMs) has been shown to significantly enhance performance \citep{kaplan2020scaling, achiam2023gpt}, augmented by alignment \citep{bai2022constitutional, rafailov2024direct, schulman2017proximal} to make LLMs follow human instructions. Instruction fine-tuning (IFT) \citep{weifinetuned, longpre2023flan, sanhmultitask, ouyang2022training} has become an essential step in adapting LLMs to perform effectively across diverse tasks, and it is believed that the data quality and diversity are the most important factors during IFT.
Recently, there is a growing interest on data-centric AI.
For example, innovations in data engineering enable scaling to vast contexts, extending the model’s capacity to handle extensive data sequences \citep{fudata}. Alpaca \citep{taori2023stanford} can elicit Llama's instruction-following by distilling $52$k instruction conversations from ChatGPT \citep{chatgpt2023}.
Importance resampling for selection has been developed to optimize data relevance \citep{xie2023data}.

Although it is well acknowledged that the critical to effective instruction tuning is the training data's quantity \citep{ding2023enhancing}, quality \citep{chenalpagasus}, diversity \citep{bukharin2023data}, and complexity \citep{wang2023self}, as also highlighted by recent industrial technical reports, such as Llama-3 \citep{dubey2024llama} and QWen-2 \citep{yang2024qwen2},
it is still a mystery of how to accurately measure those features in data.
Generally, ensuring data quantity and quality are easier to achieve with either human feedback \citep{ouyang2022training} or automated AI feedback \citep{lee2023rlaif}.
However, evaluating data diversity and complexity are challenging. For instance, a recent call \citep{zhao2024position} urge a quantifiable approach to dataset diversity to ensure that claims of diversity are substantiated and meaningful: "\textbf{Measure Dataset Diversity, Don't Just Claim It}".

Previous work has proved the effectiveness of small data quantity for achieving either general instruction following \citep{chen2023maybe, zhou2024lima}, or task-oriented abilities \citep{xialess}. \cite{zhang2024instruction} shows some preliminarily results to claim that instruction diversity drives generalization to unseen scenarios through symbolic tasks. And \cite{liumakes} perform controlled studies to measure data across
three dimensions: complexity, quality, and diversity (measured by cosine similarity of sentence representation). Besides, data pruning based on concept complexity has been shown effective in both pre-training \citep{abbaseffective} and post-training \citep{lu2023instag}.
However, many previous text diversity metrics \citep{shaib2024standardizing} are not adopted due to their intrinsic limitations. Therefore, there is still a lack of reliable diversity measure toward more efficient, effective, and targeted data selection in the instruction tuning landscape. And it is also unclear about the underlying mechanism such as what contributes to an accurate data diversity measure. 
Recently, SAEs \citep{cunningham2023sparse} has emerged as a powerful tool for interpreting the activated features in language models. The sparsity of the atomic monosemanticity guarantees the features are highly independent and 
accurate. Inspired by this advance, we first train an SAE and then \textbf{propose a new approach to use the activated features from SAE to measure data diversity}. 
Depending on the number of selected data, we then proposed two algorithms for effective data selection. The first one is greedy sampling using features from SAE for limited data (\textbf{SAE-GreedSelect}) and the second one is similarity-based sampling using features from SAE for scaling up the data selection (\textbf{SAE-SimScale}). We conduct comprehensive experiments on the widely used Alpaca \citep{taori2023alpaca} and WizardLM\_evol\_instruct\_70k \citep{xu2023wizardlm} datasets and our approach witnesses significant superiority over several previous results such as \textit{\#InsTag} \citep{lu2023instag}, \textit{Longest} \citep{zhaolong} and Repr Filter \citep{liumakes} across various models and data scale. 
Our approach also explains why some previous methods work by looking at the extracted features in the selected data.

In summary, we propose a novel method to use SAEs for diversity measure, and design new algorithms for diversity-driven data selection, contributing to more effective and interpretable instruction tuning.

\section{Related Work}

\textbf{Sparse Autoencoders} (SAEs) are powerful for understanding neural representations \citep{gao2024scaling, paulo2024automatically, braun2024identifying}, as well as in scaling and enhancing the interpretability of LLMs \citep{cunningham2023sparse, foote2023neuron}.
The foundational work in sparse coding, also known as dictionary learning, dates back to an overcomplete basis set proposed around 30 years ago \citep{olshausen1997sparse}. Based on this, K-SVD \citep{aharon2006k} was developed as an algorithm for designing overcomplete dictionaries, facilitating sparse representations in signal processing. Furthermore, K-SAEs \citep{makhzani2013k} introduced a K-sparse constraint to enforce sparsity, leading to more accurate data representations.
\cite{bau2020understanding} highlights the importance of understanding unit-specific functions for the interpretability of the model. Furthermore, \cite{tonolini2020variational} proposes that the visual cortex can employ such strategies to efficiently represent natural images.
Recently, sparse dictionary learning has been applied to visualize transformer models \citep{elhage2022toy, henighan2023superposition}, revealing that contextualized embeddings can be expressed as linear combinations of transformer factors \citep{yun2021transformer}. Research has also investigated polysemanticity and capacity in neural networks, finding that neurons often encode multiple concepts, which poses challenges for interpretability \citep{scherlis2022polysemanticity, lieberum2024gemma}. Moreover, engineering monosemanticity has been explored to design neurons that represent single concepts, thus enhancing interpretability \citep{jermyn2022engineering}.
A structured mathematical framework has been proposed for analyzing transformer circuits \citep{elhage2021mathematical}. It has been demonstrated that LLMs can elucidate the functions of neurons within themselves, offering a novel perspective on model introspection \citep{bills2023language}.
Sparse probing techniques have identified significant neurons, providing case studies that underscore the utility of sparsity in interpretability \citep{gurneefinding}. Improvements in dictionary learning with Gated-SAE \citep{rajamanoharan2024improving} and JumpReLU-SAE \citep{rajamanoharan2024jumping} further enhanced the quality of learned representations.

\textbf{Data Selection} \citep{albalak2024survey, wang2024survey}, the process of selecting a representative subset of data to achieve efficient training without compromising model performance, is important in both pre-training \citep{brandfonbrener2024color, tirumala2023d4} and post-training \citep{chenalpagasus, li2024superfiltering}.
Previous studies \citep{mindermann2022prioritized, paul2021deep} have focused on optimizing this selection for various training objectives, specifically targeting model performance constraints \citep{xia2024refined}, the value of individual data points \citep{covertscaling}, and addressing bias \citep{jain2024data}.
Recently, the emphasis on data curation \citep{taori2023stanford, chiang2023vicuna, cui2024ultrafeedback, wang2023self} and selection \citep{zhou2024lima} for LLMs suggests that the main capabilities of LLMs come from pre-training, and a small amount of well-crafted instruction data can enable excellent instruction following.
As a result, various data selection methods \citep{du2023mods, chen2023maybe, xia2024rethinking, ge2024clustering, lee2024concept, liu2024selectit} have been proposed. For example, AlpaGasus \citep{chenalpagasus} uses ChatGPT to score data quality and selects only the top 1000 highest-scoring data points. \cite{zhaolong} proposes a simple yet effective baseline of selecting the longest responses, while \cite{xialess, zhang2024tagcos, pan2024g} employ gradient-based clustering for task-agnostic coreset selection.
Further research \citep{liumakes} indicates that data quality \citep{li2024quantity, ding2023enhancing, li2024selective, li2023reflection}, diversity \citep{ge2024clustering}, and complexity \citep{xu2023wizardlm, sun2024conifer, ivison2023camels} are all crucial to the success of IFT, especially under complex constraints. However, accurately measuring these dimensions is a nontrivial task.
\textbf{In this work, we focus on data diversity}. Data diversity in instruction-tuning is increasingly recognized as crucial for building robust models \citep{bukharin2023data}. \textit{\#InsTag} \citep{lu2023instag} measures data diversity through intention tags of instructions, a method also adopted in the official technical reports of advanced LLMs, including Llama-3 \citep{dubey2024llama} and Qwen2-72B \citep{bai2023qwen, yang2024qwen2}. In this work, we prove that SAEs offer a better diversity measure.

\section{Sparse Autoencoder}\label{sec:method}

\textbf{Formulation.}
Let $x^j$ denotes the $d$-dimension residual stream of the $j$th-layer of the transformer layer before the final layer of logits (we discard $j$ hereafter and only use $x$ for simplicity). A typical SAE can be decomposed into two parts: the encoder and decoder, where the encoder is used for representation decomposition and the decoder is only used during training for loss reconstruction and discarded during inference. A ReLU-SAE can be written as 
\begin{equation}
\begin{aligned}
   \textit{Encoder: } z &= \text{ReLU}(W_{\text{encoder}}(x - b_{\text{norm}}) + b_{\text{encoder}}) \\
   \textit{Decoder: }  \hat{x} &= W_{\text{decoder}}z + b_{\text{norm}}
\end{aligned}
\end{equation}
, where $b_{\text{norm}}$ means all inputs are normalized to the unit norm before passing to the autoencoder and computing the reconstruction errors. 
Notice that $W_{\text{encoder}}$ $\in$ $\mathbb{R}^{n \times d}$, $b_{\text{encoder}}$ $\in$ $\mathbb{R}^n$, where $n$ is number of pre-defined latents in SAE.

Similarly, a k-sparse autoencoder \citep{makhzani2013k} regulates the count of active latent units by employing the TopK activation function, which retains only the k largest latent values, setting all others to zero. Thus,
\begin{equation}
\begin{aligned}
   \textit{Encoder: } z &= \text{TopK}(W_{\text{encoder}}(x - b_{\text{norm}}) )
\end{aligned}
\end{equation}
and the decoder is the same. The model is trained by gradient descent through training loss $\mathcal{L} = \|x - \hat{x}\|_2^2$. The TopK-SAE has already been verified by OpenAI \citep{gao2024scaling} for its superior performance in explaining GPT-4, thus we also follow their training tricks, such as forcing dead neuron activation if it has not activated in 10 million tokens (with an auxiliary loss coefficient of $1/32$), normalizing the decoder weights to have unit norm, and setting $W_{\text{decoder}}$ to be the transpose of $W_{\text{encoder}}$ for improving the training process.

\textbf{SAE Training.}
We pick the last residual stream from the final layer of Llama-3.1-8b-instruct \footnote{For better visualization, we summarize all the used open-source links in App. \ref{app:links} hereafter.} for training our TopK-SAE. And we set the number of latents $n$ to $131,072$,  and $K$ to \{16, 32, 64, 128, 256\}. We use Top-$128$-SAE for our main experiments for its moderate size. 
We use the 10B tokens from RedPajama-Data-V2 \citep{together2023redpajama} for SAE training considering its high quality and diverse source. We tried various batch sizes from \{$4,096$, $8,192$, $12,288$\} and found $4,096$ to be the optimal, and more can be found in Appendix \ref{app: sae-bs}. 
For a total batch size of $4,096$, the batch size is 32 per device, and grad\_acc\_steps and micro\_acc\_steps are 4 and 2, respectively.
For all training, we use 4 nodes with 8 Nvidia A100 80G GPUs per node through model parallel. The lr\_warmup\_ratio is $0.5$ and the learning rate is $7e-5$. 
We set epoch to 4 and we do not find additional benefits with more epochs. 
Since the Llama-3 tokenizer contains a $BOS$ token which is useless and even detrimental to SAE training, we discard all $BOS$ tokens for SAE training. We preprocess all the data to the same length through concatenating and chunking passages.

\begin{figure}[ht]
\centering
\includegraphics[width=0.95\linewidth]{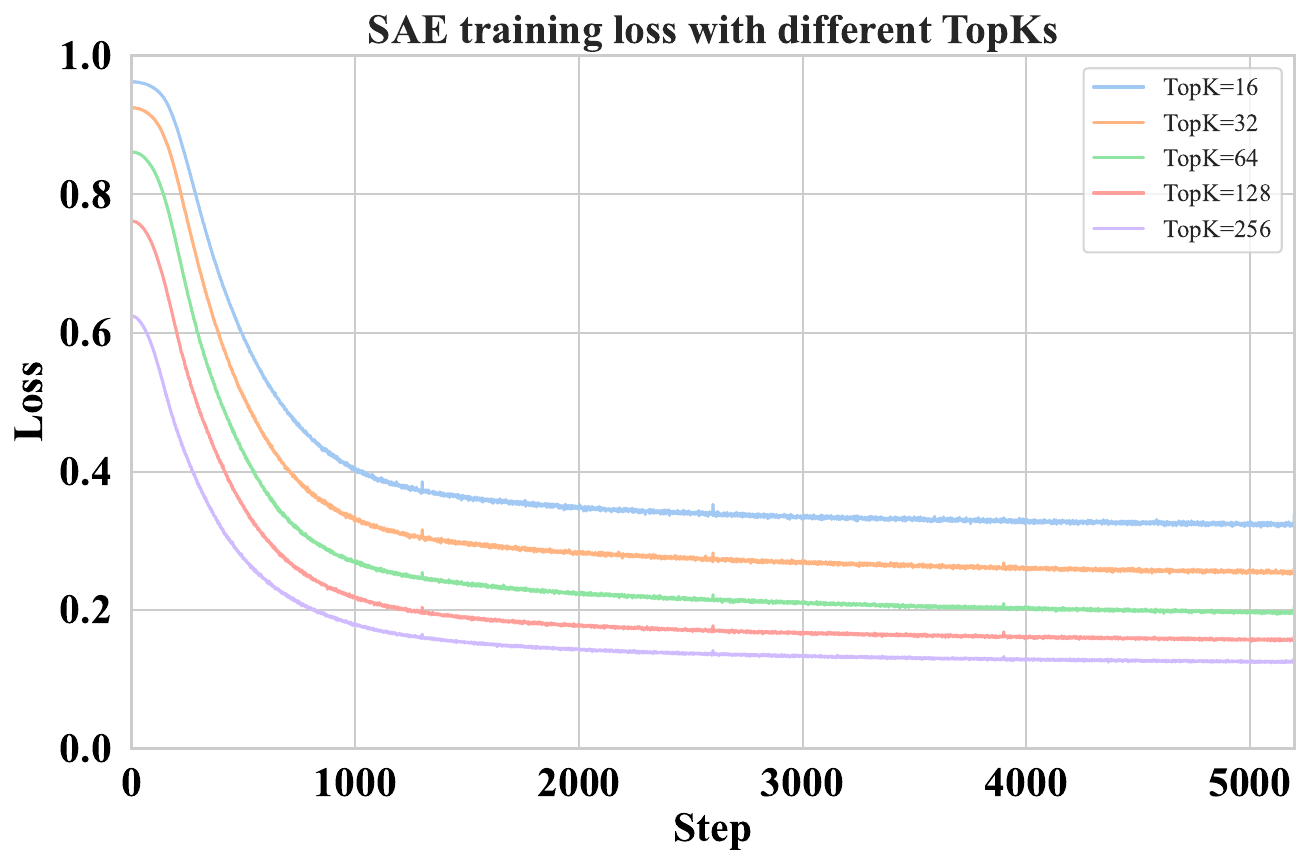}
\caption{The training loss of TopK-SAE on the layer 31 of Llama-3.1-8b-instruct.}\label{fig: loss_sae31}
\end{figure}

Figure \ref{fig: loss_sae31} shows the training loss over different TopKs under the same configurations. It is generally observed that the training loss almost saturates after $4,000$ steps, and a larger k leads to a better final loss. However, a smaller loss does not necessarily correlate with better-decomposed features since more activated features will be more difficult to interpret. 

\textbf{SAE Inference}
After training the SAEs, inspired by the significant performance boost achieved with JumpReLU-SAE \citep{rajamanoharan2024jumping}, we further use JumpReLU during inference to rectify the activations and only treat the activation value larger than the threshold as true activation. 

\begin{equation}
\text{JumpReLU}(x) =
\begin{cases} 
x, & \text{if } x > \theta, \\
0, & \text{otherwise.}
\end{cases}
\end{equation}
, where $\theta$ is the jump threshold.

\begin{table*}[htbp]
    \centering
    \scalebox{0.86}{
    \begin{tabular}{c|c|cccc|cccc}
        \toprule
        \multirow{2}{*}{Data} &\multirow{2}{*}{Method} & \multicolumn{4}{c|}{\textbf{Alpaca-52k}} & \multicolumn{4}{c}{\textbf{WizardLM\_evol\_instruct-70k}} \\ \cline{3-10}
        & & \tiny{\makecell{\textbf{Strict} \\ \textbf{prompt-level}}} & \tiny{\makecell{\textbf{Strict} \\ \textbf{instruction-leve}}} & \tiny{\makecell{\textbf{Loose} \\ \textbf{prompt-level}}} & \tiny{\makecell{\textbf{Loose} \\ \textbf{instruction-level}}} & \tiny{\makecell{\textbf{Strict} \\ \textbf{prompt-level}}} & \tiny{\makecell{\textbf{Strict} \\ \textbf{instruction-leve}}} & \tiny{\makecell{\textbf{Loose} \\ \textbf{prompt-level}}} & \tiny{\makecell{\textbf{Loose} \\ \textbf{instruction-level}}} \\ \cline{1-10}
        All & NA & 36.23 & 46.64 & 37.89 & 48.68 & 35.49  & 46.76   & 40.85  & 52.04 \\ \cline{1-10}
        \multirow{6}{*}{1k} 
        &\#INSTAG \citep{lu2023instag} & 20.70 & 29.38 & 21.26 & 30.34 & {25.51}  & {36.81}  & \textbf{29.57} & {41.13} \\ 
        &Longest-instruction & 18.30  & 29.38  &  23.48 & 35.37  & 23.48 & 34.53 & 26.62 & 38.61 \\  
        &Longest-response \citep{zhaolong} & 19.41 & 30.09 & 22.92 & 34.53 & 22.37 & 34.89 & 27.36 & 40.29 \\ 
        &Repr Filter \citep{liumakes} &  21.63 & 31.41 & 23.29  & 32.97   & 24.95 & 35.73 & 27.17 & 38.25 \\  \cline{2-10}
        &\textbf{\one} & \textbf{26.80} & \textbf{38.61} & \textbf{27.91} & \textbf{44.89} & {24.77} & {35.97} & {28.47} & {40.05} \\ \cline{2-10} 
        &\textbf{\two} & {23.48} & {34.65} & {24.77} & {37.41} & \textbf{25.80} & \textbf{38.69} & 29.21 & \textbf{41.33} \\ 
        \midrule
        \multirow{6}{*}{3k}
        &\#INSTAG \citep{lu2023instag} & 22.37 & 33.57 & 23.48 & 35.25 &30.87  & {41.97}  & {34.38} & {46.16} \\  
        &Longest-instruction & \textbf{30.31}  & 40.41  & 31.98  & 43.29  & 28.28  & 40.17  & 33.09  & 45.08  \\
        &Longest-response \citep{zhaolong} & 21.81 & 33.69 & 24.95 & 37.05 & 24.95 & 37.53 & 31.61 & 43.53 \\ 
        &Repr Filter \citep{liumakes} & 23.11 & 34.29  & 24.21 & 36.45  & 28.47 & 40.17 & 31.42 & 44.48 \\  \cline{2-10}
        &\textbf{\one} & 29.76 & \textbf{40.89} & \textbf{32.16} & \textbf{43.65} &  \textbf{34.75} &  \textbf{47.36} &  \textbf{38.08} &  \textbf{50.96} \\ \cline{2-10}
        &\textbf{\two} & 28.84 & 40.41 & 31.05 & 42.93 & 29.76 & 40.41 & 33.83 & 44.24 \\ 
        \midrule
        \multirow{6}{*}{5k}
        &\#INSTAG \citep{lu2023instag} &  NA & NA  & NA  & NA & 32.72  & 43.28  & 36.41 & 47.36 \\  
        &Longest-instruction & 27.36  &  39.09 &  29.76 & 41.97  & 31.61  & 42.45  & 36.04  & 47.72  \\
        &Longest-response \citep{zhaolong} & 21.99 & 33.69 & 24.77 & 36.81 & 27.54 & 39.69 & 33.09 & 44.96 \\ 
        &Repr Filter \citep{liumakes} & 27.17  & 36.93  & 27.91 & 37.89  & 30.50 & 42.81 & 33.46 & 46.40 \\  \cline{2-10}
        &\textbf{\one} & 26.99 & 37.41 & 27.73 & 38.97 & 31.98 & 43.29 & 36.04 & 48.20 \\ \cline{2-10}
        &\textbf{\two} & \textbf{29.02} & \textbf{41.13} & \textbf{31.05} & \textbf{43.65} & \textbf{33.46} & \textbf{44.72} & \textbf{36.97} & \textbf{48.44} \\ 
        \bottomrule
    \end{tabular}
    }
    \caption{Comparison of the fine-tuned Llama-2-13b's accuracy using different methods with 1k, 3k, and 5k data selected from the Alpaca and WizardLM\_evol\_instruct\_70k datasets. '-' indicates not applicable.}\label{tab: overall}
\end{table*}

\begin{algorithm}[h]
\caption{Two Diversity-driven Data Selection Methods}\label{algorithm:merged}
\begin{algorithmic}[1]
\REQUIRE The Whole Dataset $\mathcal{D}$, Sub-Dataset Size $N$, Sampling Mode (\textit{Greedy} or \textit{Similarity-based}), Threshold $\theta$ (if \textit{Similarity-based})
\ENSURE The Sampled Sub-Dataset $\mathcal{D}_s$
\STATE Initialize Empty $\mathcal{D}_s$
\STATE Sort Queries in $\mathcal{D}$ by instruction length in descending order
\WHILE{$|\mathcal{D}_s| < N$}
    \STATE Set $\mathcal{T}_s^B \leftarrow \emptyset$  \#$\mathcal{T}$ is the set of activated features
    \FOR{each Query $q \in \mathcal{D}$}
        \IF{\textcolor{blue}{$|\mathcal{T}_s^B \cup \mathcal{T}_q| > |\mathcal{T}_s^B|$ \textbf{(if \textit{Greedy: \one})} } \textbf{or} 
        \textcolor{red}{ $|\mathcal{T}_s^B \cap \mathcal{T}_q| / |\mathcal{T}_s^B| < \theta$ \textbf{(if \textit{Similarity-based: \two})}}}
            \STATE $\mathcal{D}_s \leftarrow \mathcal{D}_s \cup \{q\}$
            \STATE $\mathcal{T}_s^B \leftarrow \mathcal{T}_s^B \cup \mathcal{T}_q$
            \STATE $\mathcal{D} \leftarrow \mathcal{D} \setminus \{q\}$
            \IF{$|\mathcal{D}_s| = N$}
                \STATE break
            \ENDIF
        \ENDIF
    \ENDFOR
\ENDWHILE
\STATE \textbf{return} $\mathcal{D}_s$
\end{algorithmic}
\end{algorithm}

\section{Methods and Experiments} 

\subsection{Diversity-driven Data Selection through Sparse Autoencoder}
We only focus on \textbf{diversity} for the selection to prove the effectiveness of SAEs against existing diversity measures.
Based on the extracted features from our trained SAEs, we design two data selection methods: 
1) when we only want to select a limited number of data as presented in \citep{chenalpagasus, zhaolong}, we propose greedy sampling using features from SAE for limited data (\textcolor{blue}{\textbf{SAE-GreedSelect}}) to maximize the utilization of features, and 
2) when we want to scale up the selected data rather than only picking a fixed number of data, we propose similarity-based sampling using features from SAE for scaling up the data selection (\textcolor{red}{\textbf{SAE-SimScale}}). For example, the original \#INSTAG \citep{lu2023instag} method uses a greedy search but can not scale to $5$k data since no data can bring new intention tags. 
The details of the two methods can be found in Algorithm \ref{algorithm:merged}.
Compared with previous methods used in the industrial model training pipeline \citep{lu2023instag}, our method is designed to be \textbf{simple and scalable so that it can also be used for large-scale training by the industry}.

As we mentioned above that we actually use JumpRelu during inference, and we empirically set the threshold to $10$ across all experiments in our work. For method \textbf{SAE-SimScale}, there is one additional parameter of similarity ratio, and we set it to $0.8$ for all experiments. Thus, there are only $1$ and $2$ tunable parameters for \textbf{SAE-GreedSelect} and \textbf{SAE-SimScale}, respectively. This way, we can maximize the simplicity.
Additional experiments regarding the threshold can be found in Section \ref{sec: threshold}, and some even witness better results.

\subsection{Experimental Settings}

\textbf{Models.} To validate our method for data selection for supervised instruction fine-tuning (SFT), we use Llama-2-13b-base for our main experiments as the foundation base model. In addition, we also use Gemma-2-9b and Llama-2-7b-base to verify the method at different model scales.

\textbf{Datasets.} We use Alpaca-52k and WizardLM\_evol\_instruct\_70k as the target instruction tuning datasets because they are widely used and contain large enough data points for data selection.
 
\textbf{Baselines.} We choose Longest-response \citep{zhaolong}, \textit{\#Instag} \citep{lu2023instag} and Repr Filter \citep{liumakes}  as the baselines. 
\textit{\#Instag} \citep{lu2023instag} and Repr Filter \citep{liumakes} share similar ideas of gradually selecting dadapoints that add additional diversity, measured by intention tags and Cos similarity (we follow the similarity threshold of $0.9$ in their paper but replace the sentence embedding with Llama-3-8b), respectively.
Additionally, we also use the longest instruction as a baseline for comparison with our methods. 
Note that the best performance in longest-response \citep{zhaolong} was achieved by additional tricks such as refining the instructions or NEFTune \citep{jainneftune}, but we did not follow this for a simple and fair comparison with other baselines.

\textbf{Configurations.} For all SFT experiments, we use 8 Nvidia A100 80G GPUs and the stanford alpaca code base. Following similar configurations in \citep{zhaolong}, we set the maximum length to be $1,024$ for data selected from Alpaca-52k and $2,048$ from WizardLM\_evol\_instruct\_70k. We always set the batch size to 128, learning\_rate to 1e-5, weight\_decay to 0.1, and warmup\_ratio to 0.03. We set epochs to 15 for $1$k and $3$k data, $5$ for $5$k data, and $3$ for full data.

\textbf{Evaluation.}
The selection of SFT data aims to elicit superior instruction-following abilities, but the evaluation lacks standardization.
For a comprehensive evaluation, we use IFEval \citep{zhou2023instruction} for strict evaluation since it can accurately measure the response that adheres to the complex instructions through verifiable instructions, such as  "mention the keyword of AI at least 3 times".
We also use LLM- and Human-as-a-Judge for head-to-head evaluation and report the results on the AlpacaEval 2.0 leadboard as additional metric.
Besides, it is expected that the models trained on small instruction datasets also behave well in other knowledge-intensive tasks, such as MMLU \citep{hendrycksmeasuring}, TruthfulQA \citep{lin2022truthfulqa}, Winogrande \citep{sakaguchi2020winogrande}, Arc \citep{clark2018think} and Gsm8k \citep{cobbe2021training}. So we also use those five benchmarks for measuring the finetuned model's knowledge.
For all the evaluations on those five benchmarks, we utilize the lm-evaluation-harness under their default setting with the same number of $3$ in-context examples for all models.

\begin{figure}[h]
\centering
\includegraphics[width=0.7\linewidth]{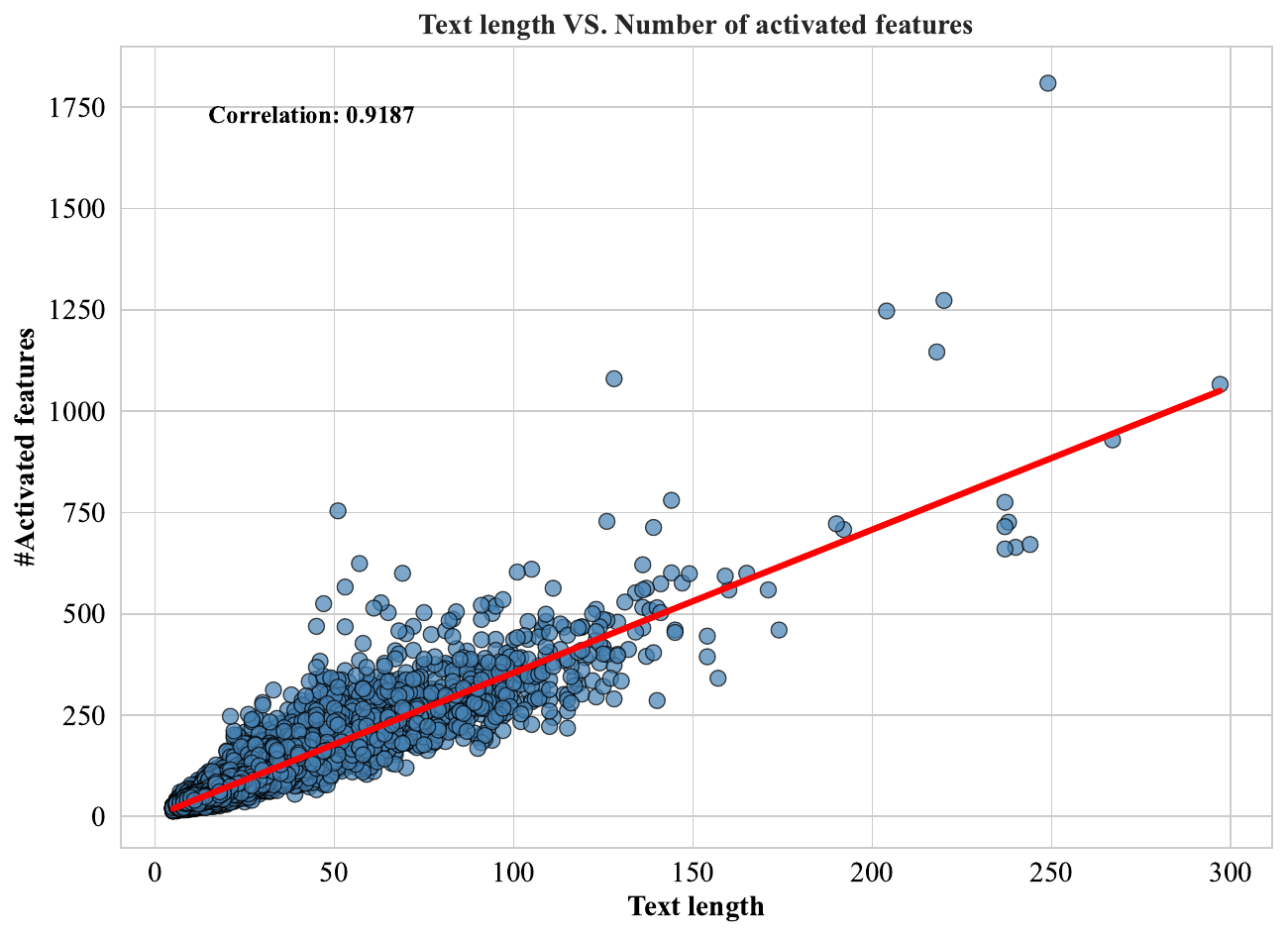}
\caption{The correlation between text length and number of activations in SAEs.}\label{fig: correlation}
\end{figure} 

\begin{figure*}[ht]
\centering
\includegraphics[width=0.9\linewidth]{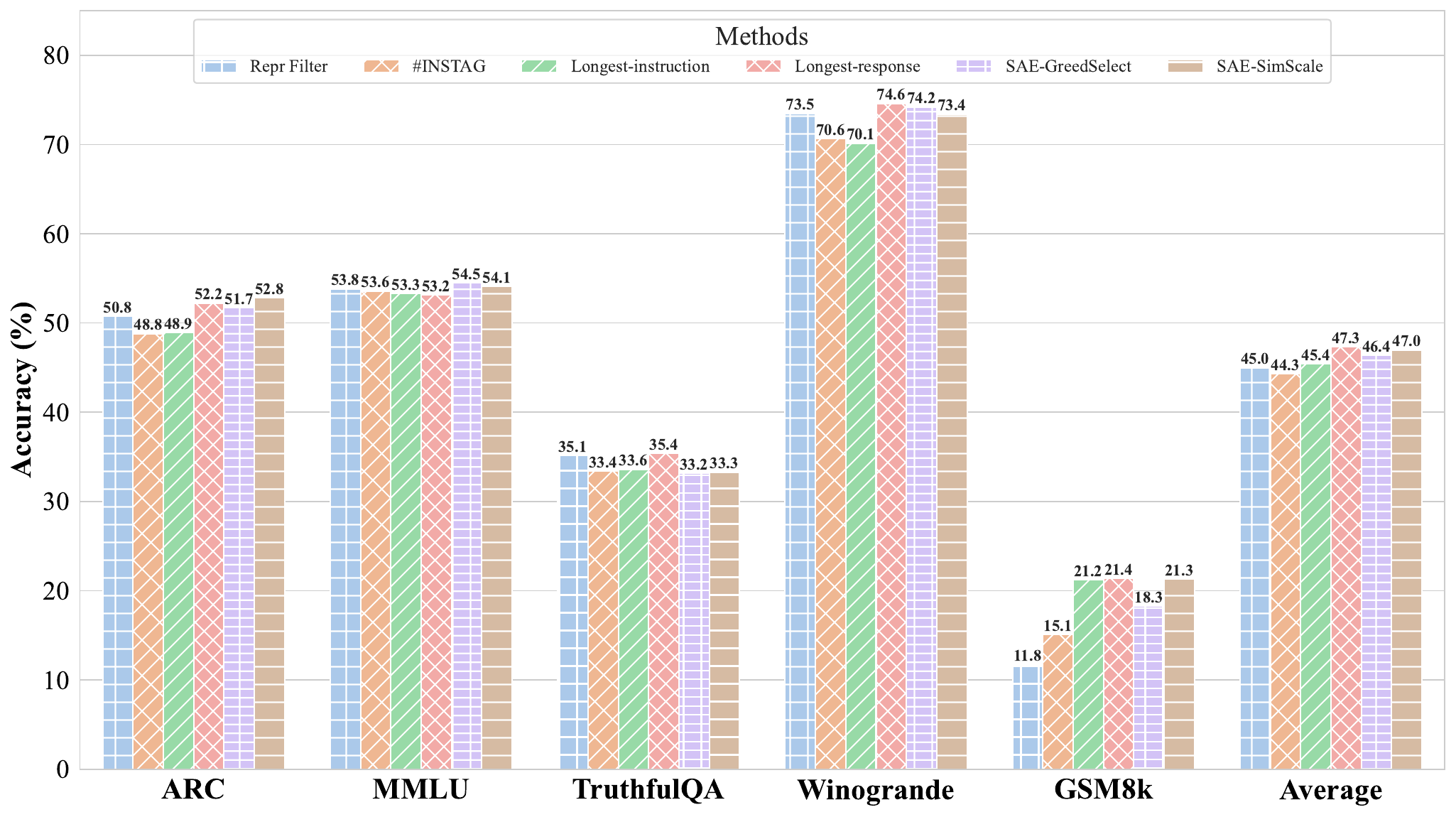}
\caption{The benchmark performance between different methods: Llama 2 (13B) trained from corresponding $1$k selected data from Alpaca.}\label{fig: benchmark_wizardlm-llama2-13b}
\end{figure*}

\section{Results and Analysis}
\subsection{Why $1,000$ Longest Responses \citep{zhaolong} Lead to Strong Performance?}
Picking the $1,000$ longest responses \citep{zhaolong} serves as a simple but tough-to-beat baseline for SFT and they hypothesize that this is because the longest responses intuitively contain more learnable information, but do not provide quantitative evidence to this assumption. Thus, we provide a quantitative explanation of the activated features through SAE by examining the correlation between activated features and text length.
Figure \ref{fig: correlation} shows a strong positive correlation (r = $0.92$) between text length and feature richness in an SAE extracted from text in the Alpaca dataset, supporting their hypothesis that longer responses encode more learnable information. This simple strategy provides a competitive baseline for fine-tuning LLMs efficiently but is not always better, as shown later.

\subsection{Comparison with Baselines}

Table \ref{tab: overall} presents a comparative analysis of different data selection methods applied to Alpaca-52k and WizardLM\_evol\_instruct-70k datasets, evaluated under "Strict prompt-level," "Strict instruction-level," "Loose prompt-level," and "Loose instruction-level" on the IFEval.
As the data size increases from $1$k to $5$k, the performance of all methods improves across both datasets.
Our proposed methods \one and \two exhibit more significant gains than baseline approaches, indicating their robustness and scalability. For instance, in the WizardLM\_evol\_instruct-70k dataset at the 3k data scale, \two achieves a "Loose instruction-level" score of 50.96, significantly surpassing the \#instag baseline's 46.16 and other baselines. And it even matches the upper bound of using the entire $70$k data.
\two achieves the best results overall, particularly with higher data sizes ($3$k and $5$k), with significant performance improvements in "Strict instruction-level" and "Loose instruction-level" evaluations. Performance differences are more pronounced in the WizardLM\_evol\_instruct-70k dataset, highlighting the challenges and opportunities in leveraging larger and more complex instruction-tuning datasets.

The results suggest that the \two method is the most effective approach for data selection through diversity-driven frameworks.
By outperforming baselines across all metrics and datasets, \two highlights its potential for optimizing \textbf{scalable data selection}.

\begin{figure*}[ht]
\centering
\includegraphics[width=0.7\linewidth]{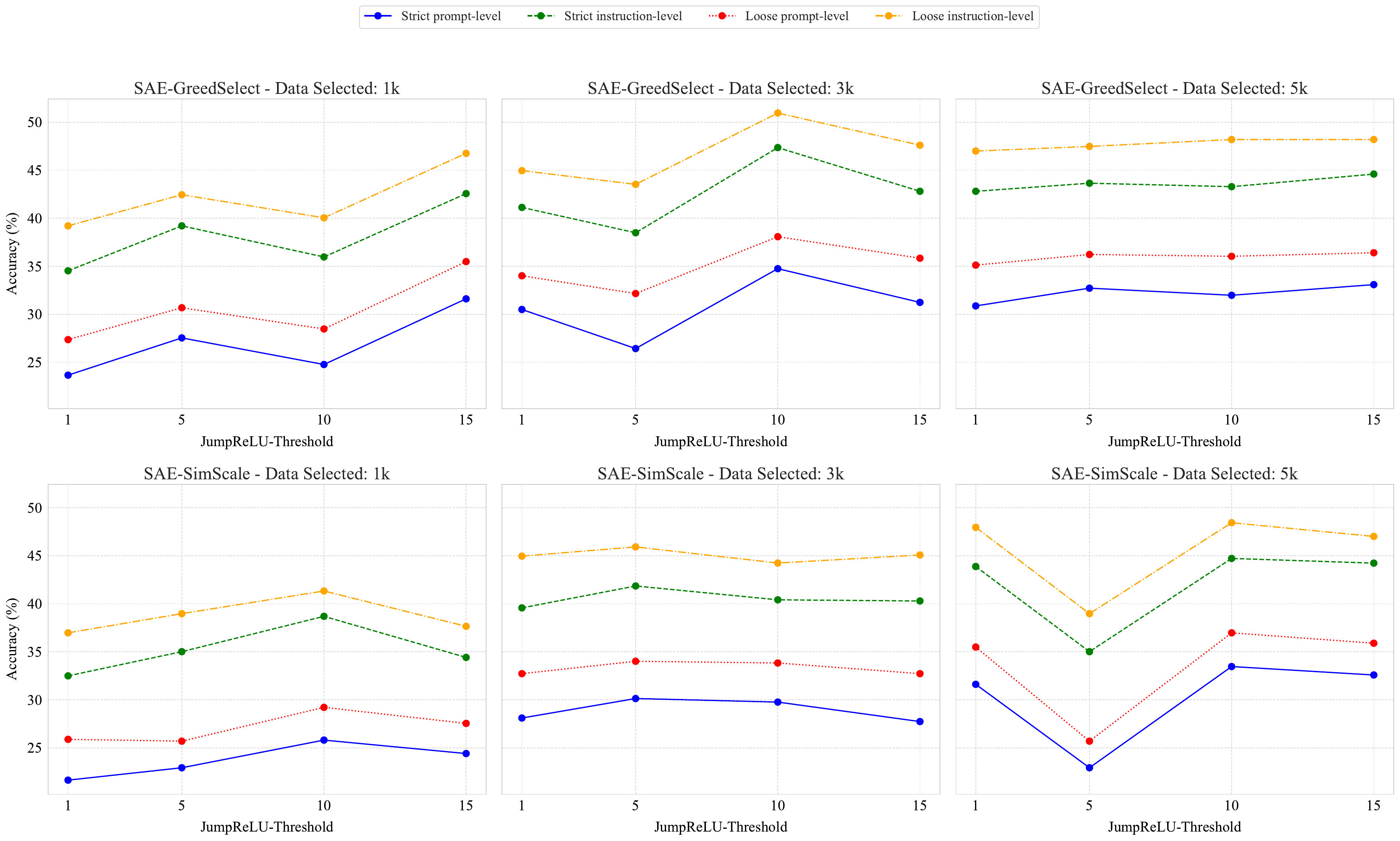}
\caption{The comparison of \one (\textbf{Top}) and \two (\textbf{Bottom}) under different SAE thresholds.}\label{fig: threshold_wizard}
\end{figure*}

\begin{table}[htbp]
    \centering
    \scalebox{0.77}{
    \begin{tabular}{c|c|cccc}
        \toprule
        \multirow{2}{*}{Data} &\multirow{2}{*}{Method} & \multicolumn{4}{c}{\textbf{Alpaca-52k}} \\ \cline{3-6}
        & & \tiny{\makecell{\textbf{Strict} \\ \textbf{prompt-level}}} & \tiny{\makecell{\textbf{Strict} \\ \textbf{instruction-leve}}} & \tiny{\makecell{\textbf{Loose} \\ \textbf{prompt-level}}} & \tiny{\makecell{\textbf{Loose} \\ \textbf{instruction-level}}}   \\ \cline{1-6}
        \multirow{2}{*}{1k}
        &\textbf{Gemma-SAE} & 23.66 & 34.05 & 24.58 & 35.25    \\ 
        &\textbf{Ours-SAE} & 24.21 & 35.13 & 26.06 & 37.53 \\ 
        \midrule
        \multirow{2}{*}{3k}
        &\textbf{Gemma-SAE} & 31.61 & 41.96 & 33.27 & 43.88   \\ 
        &\textbf{Ours-SAE} & 29.76 & 40.89 & 32.16 & 43.65  \\ 
        \bottomrule
    \end{tabular}
    }
    \caption{Comparison of different methods using 1k data from Alpaca datasets. }\label{tab: different_sae}
\end{table}

\subsection{Preference Evaluation}
\begin{figure}[h]
\centering
\includegraphics[width=0.8\linewidth]{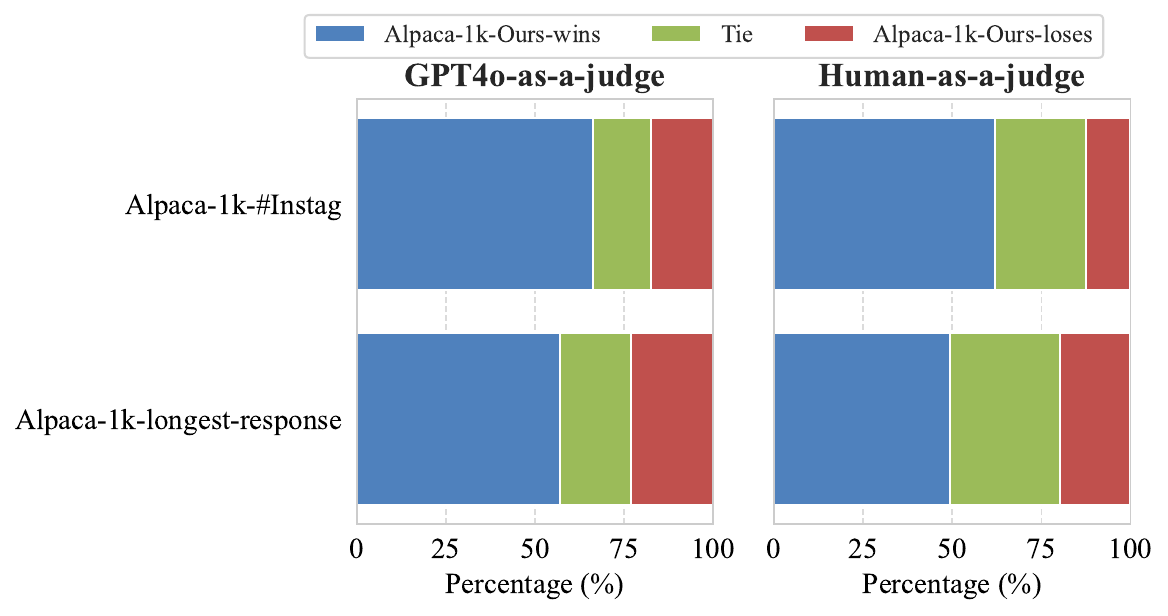}
\caption{The head-to-head comparison between Llama-2-13b-base trained on different data selected from Alpaca.}\label{fig: headtohead1}
\end{figure}

In addition to the above evaluation, we also focus on assessing human or AI preference evaluation. The head-to-head comparison is shown in Figure \ref{fig: headtohead1}, showing that our methods again consistently outperform the baselines. More details can be found in the Appendix \ref{app: comparison}.

Besides, In Table \ref{tab: leadboard} we report the results on the AlpacaEval 2.0 benchmark with some baselines copied from their public leaderboard and \citep{zhaolong}. The bottom four models are all trained on Llama-2-13b-base with $5$k data selected from the WizardLM\_evol\_instruct-70k dataset. Notably, our method achieves superior performance compared to some commercial models, including ChatGPT and Claude, outperforms the baseline approaches by a large margin. 

\begin{table}[h]
    \centering
    \scalebox{0.9}{
    \begin{tabular}{lcc}
        \toprule
        \small{Model} & \small{Win Rate}  & \small{Length Controlled Win Rate} \\
        \midrule
        \small{Wizardlm-70b} & 14.38   &  17.57  \\
        \small{AlpaGasus-1k} &   4.87 & -  \\
        \small{LIMA-1k} &  5.64  &  -  \\
        \small{gpt-3.5-turbo-1106} &  9.18  &  19.3  \\
        \small{claude-2.1\_concise} &  9.23  & 18.21   \\
        \midrule
        \small{\#INSTAG-5k} &  3.42 &  6.08   \\
        \small{Longest-response-5k} & 5.86 &  7.04  \\
        \small{Repr Filter} &  4.91&  9.55   \\
        \small{\textbf{SAE-GreedSelect-5k}} &  \textbf{19.61} &  \textbf{22.81}   \\
        \bottomrule
    \end{tabular}
    }
    \caption{Preference evaluation results on AlpacaEval 2.0. '-' represents not available.}\label{tab: leadboard}
\end{table}

\begin{figure}[h]
\centering
\includegraphics[width=0.7\linewidth]{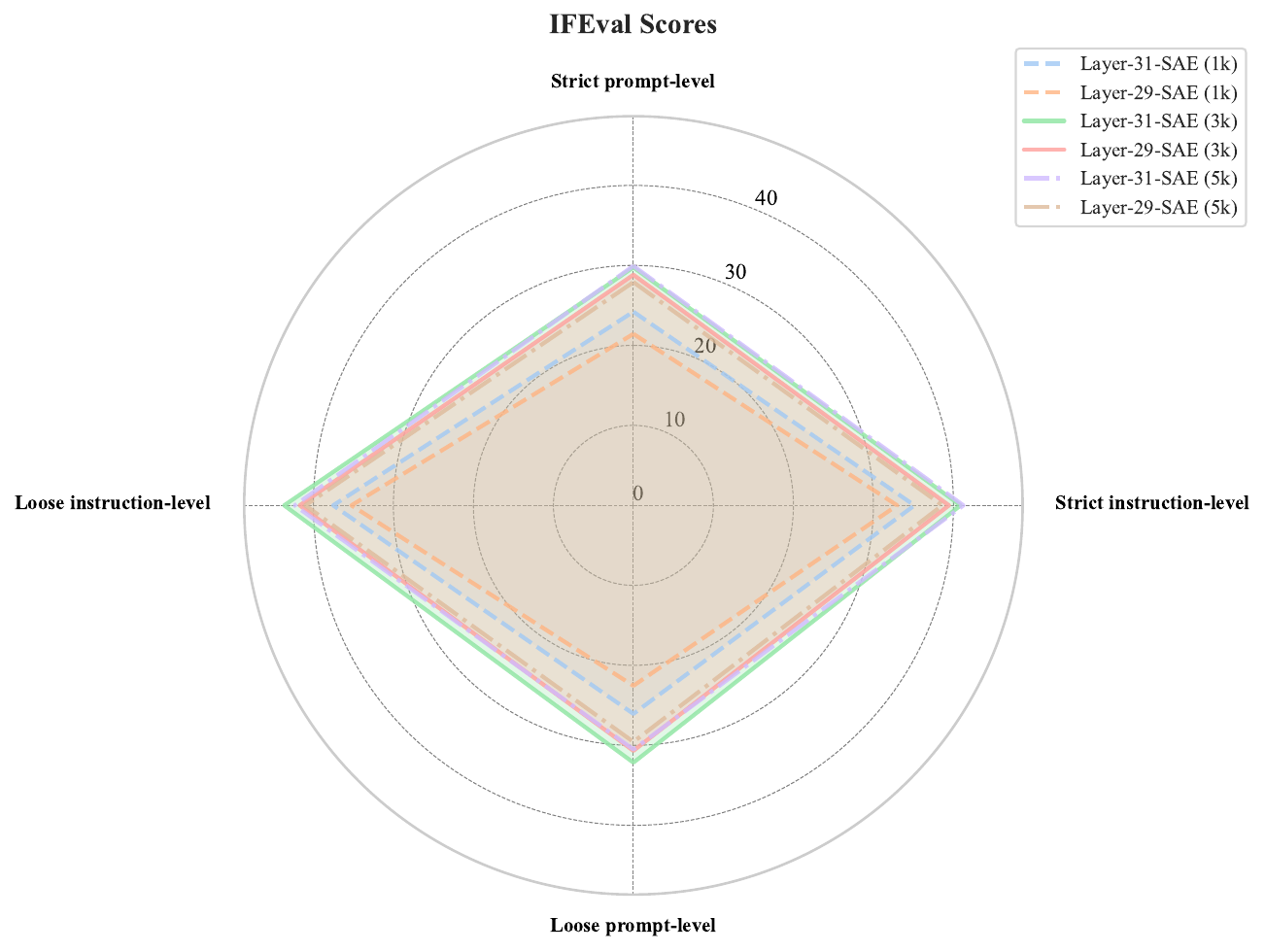}
\caption{Radar plot comparing the results of using SAEs on different layers.}\label{fig: different_layer}
\end{figure} 

\begin{figure}[h]
\centering
\includegraphics[width=0.95\linewidth]{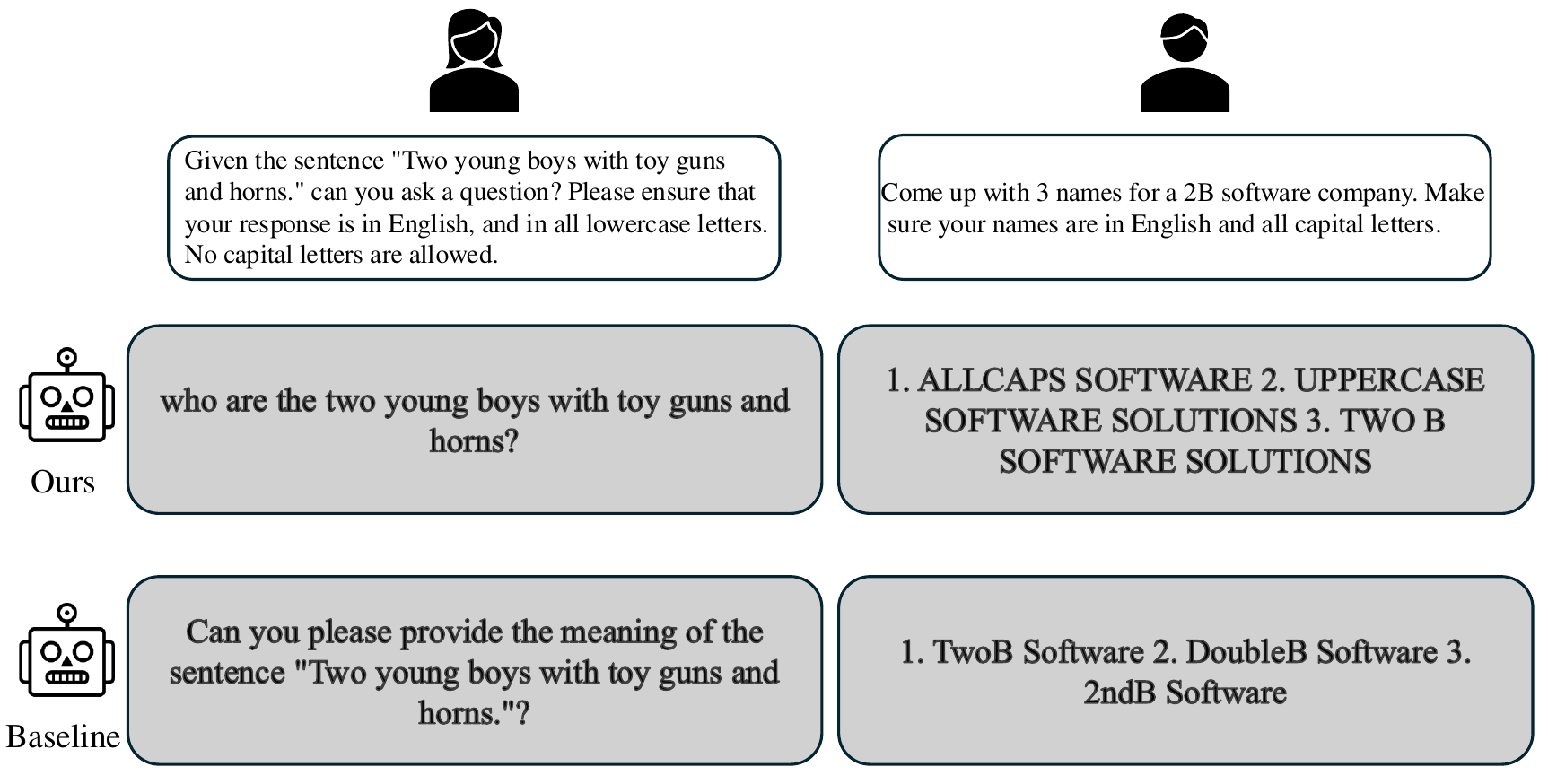}
\caption{Case study: the
instruction-following performance of Llama-2-13B model finetuned
on $3$k data selected from Alpaca with our \one and \textit{\#INSTAG} baseline.}\label{fig: case_study}
\end{figure}

\subsection{Different SAE as Feature Extractor}

In Table \ref{tab: different_sae}, we show the result using different SAE as the feature extractor.
This table compares the performance of the same algorithm \one using two SAEs as the backbone, Gemma-SAE and Ours-SAE, across four evaluation metrics on subsets of the Alpaca-52k dataset with $1$k and $3$k data points. 
For the $1$k subset, Ours-SAE achieves higher accuracy scores than Gemma-SAE, such as 24.21\% vs. 23.66\% on strict prompt-level and 37.53\% vs. 35.25\% on loose instruction-level metrics. For the $3$k subset, Gemma-SAE slightly outperforms Ours-SAE in strict prompt-level (31.61\% vs. 29.76\%) and instruction-level (41.96\% vs. 40.89\%), though both methods are comparable in loose-level evaluations.
This confirms the universal effectiveness of our proposed data selection method when using different SAEs as the backbone, showing the flexibility and compatibility of our data selection strategy.

\subsection{Results on Base Model with Different Size}

\begin{table*}[htbp]
    \centering
    \scalebox{0.9}{
    \begin{tabular}{c|c|cccc|cccc}
        \toprule
        \multirow{2}{*}{\small{Data}} &\multirow{2}{*}{\small{Method}} & \multicolumn{4}{c|}{\textbf{\small{Gemma-2-9b-base}}} & \multicolumn{4}{c}{\textbf{\small{Llama-2-7b-base}}} \\ \cline{3-10}
        & & \tiny{\makecell{\textbf{Strict} \\ \textbf{prompt-level}}} & \tiny{\makecell{\textbf{Strict} \\ \textbf{instruction-leve}}} & \tiny{\makecell{\textbf{Loose} \\ \textbf{prompt-level}}} & \tiny{\makecell{\textbf{Loose} \\ \textbf{instruction-level}}} & \tiny{\makecell{\textbf{Strict} \\ \textbf{prompt-level}}} & \tiny{\makecell{\textbf{Strict} \\ \textbf{instruction-leve}}} & \tiny{\makecell{\textbf{Loose} \\ \textbf{prompt-level}}} & \tiny{\makecell{\textbf{Loose} \\ \textbf{instruction-level}}} \\ \cline{1-10}
        \multirow{2}{*}{1k} 
        &\small{Longest-instruction} & 21.07 & 30.70 & 21.63 & 32.25 & 19.78 & 29.26 & 21.07 & 31.41 \\  \cline{2-10}
        &\small{Longest-response} & 21.26 & 31.41 & 25.88 & 35.61 & \textbf{20.33} & 29.74 & \textbf{22.55} & 33.57 \\  \cline{2-10}
        &\textbf{\small{\one}} & 23.48 & 35.61 & 25.51 & 37.65 & 18.85 & 29.62 & 19.96 & 30.94 \\  \cline{2-10}
        &\textbf{\small{\two}} & \textbf{25.88} & \textbf{36.45} & \textbf{26.80} & \textbf{38.01} & 19.96 & \textbf{31.54} & 21.44 & \textbf{33.81} \\  
        \midrule
        \multirow{2}{*}{3k} 
        &\small{Longest-instruction} & 22.55 & 33.45 & 25.32 & 36.69 & 22.92 & 33.45 & 24.58 & 36.21 \\  \cline{2-10}
        &\small{Longest-response} & 23.11 & 33.93 & 25.88 & 37.05 & 22.18 & 32.37 & 24.58 & 35.49 \\  \cline{2-10}
        &\textbf{\small{\one}} & 22.55 & 34.41 & 23.84 & 35.97 & \textbf{23.11} & \textbf{35.61} & 24.21 & \textbf{37.77} \\  \cline{2-10}
        &\textbf{\small{\two}} & \textbf{25.51} & \textbf{37.41} & \textbf{26.43} & \textbf{38.49} & 22.74 & 34.17 & 24.95 & 36.45 \\ 
        \midrule
        \multirow{2}{*}{5k} 
        &\small{Longest-instruction} & 22.03 & 32.57 & 25.12 & 36.03 & 22.92 & 34.53 & 23.84 & 35.85 \\  \cline{2-10}
        &\small{Longest-response} & 21.99 & 32.13 & 25.14 & 35.37 & 21.63 & 32.25 & 24.58 & 35.37 \\  \cline{2-10}
        &\textbf{\small{\one}} & 21.34 & 32.73 & 22.55 & 34.17 & \textbf{25.51} & \textbf{35.97} & \textbf{26.43} & \textbf{37.41} \\  \cline{2-10}
        &\textbf{\small{\two}} & \textbf{24.58} & \textbf{36.09} & \textbf{26.25} & \textbf{37.89} & 24.95 & 34.53 & 26.06 & 35.85 \\
        \bottomrule
    \end{tabular}
    }
    \caption{Comparison of different base models with different scale using 1k, 3k and 5k data from Alpaca.}\label{tab: different_model}
\end{table*}
 
In Table \ref{tab: different_model}, we show the results using base models at different sizes to explore the impact of model parameters.
The results demonstrate that our proposed data selection method consistently outperforms baselines across all data scales ($1$k, $3$k, and $5$k) for both Gemma-2-9b and Llama-2-7b models.
And \two achieves the highest scores across almost all metrics, showcasing its effectiveness in improving performance over other methods.
The improvements with the proposed method are more evident in the Gemma-2-9b model compared to the Llama-2-7b, and the same trend is also observed in previous results on Llama-2-13b. 
The significance of the improvements achieved by the proposed data selection method diminishes when evaluated on smaller models, highlighting the dependence on the model scale for pronounced benefits. But this demonstrates significant importance of our method since it meets the scaling trend of current LLMs.

\subsection{Results of Using SAEs on Different Layers.}
The previous results come from picking the data using the SAE trained on the final ($31st$) layer. However, it is also possible to use other layers. Thus, we also pick the SAE trained on the $29th$ layer as the backbone and perform a similar pipeline of selecting the data and training the model.
Figure \ref{fig: different_layer} shows the radar plot to compare the four IFEval scores on models trained on $1$k, $3$k, and $5$k data selected using different SAE layers. 
For the 1k dataset, Layer-31-SAE scores ranged from $24.21$ to $37.53$, while Layer-$29$-SAE ranged from $21.44$ to $35.25$. With larger datasets ($3$k and $5$k), scores improved, with Layer-$31$-SAE reaching up to $43.65$ ($3$k) and $42.33$ ($5$k), consistently outperforming Layer-$29$-SAE, which peaked at $41.72$ ($3$k) and $41.01$ ($5$k).

\subsection{Impact of Inference Threshold of SAE}\label{sec: threshold}
In previous experiments, we always set the threshold to $10.0$ during the SAE inference. Here we show the results of using different threshold.
The figure \ref{fig: threshold_wizard} compares accuracy levels across varying thresholds for two versions of our methods ("\one" and "\two") under different data selection conditions ($1$k, $3$k, $5$k). Both methods exhibit an upward trend in accuracy as thresholds increase, although the rate of improvement varies. At lower thresholds, accuracy differences between the versions are minimal, but \two consistently outperforms \one at higher thresholds across all data sizes. Data size influences the trends: larger data size (e.g., $5$k) yields higher overall accuracy for both versions. Strict and loose prompt-level and instruction-level conditions impact the algorithms, with stricter conditions showing relatively lower accuracy compared to looser configurations. This figure demonstrates that an x-axis threshold of $10$ generally achieves the best accuracy results across all metrics and settings.

\subsection{Case Study}
Here we show some examples in Figure \ref{fig: case_study} as a case study, and the instructions are picked from the IFEval dataset. Our model outperforms the baseline in instruction adherence and generation correctness, with additional cases in App. \ref{app: case_study}.
\section{Factual Knowledge benchmarks}\label{sec:mitigation}

Despite the above instruction-following abilities, we look at the model performance on knowledge-intensive benchmarks, as shown in Figure \ref{fig: benchmark_wizardlm-llama2-13b}.
This figure presents a comparative analysis of model performance across five standard benchmarks and the average performance. It highlights that \two generally outperforms other models across most datasets, particularly in Winogrande and the overall score. GSM8k witnesses the lowest results among all datasets for all models, while Winogrande shows the highest accuracy, showcasing the variability in task difficulty.
However, there are \textbf{marginal differences between different selection methods}, suggesting that factual knowledge is almost maintained across five models, which is also proven in \citep{zhaolong} that small scale SFT does not impact abilities on knowledge benchmarks.
More can be found in Appendix \ref{tab: benchmark}.

\section{Conclusion}
 
Our findings unveil a novel measure of data diversity through the learned monosemanticity in sparse autoencoders. Based on the activated features in the SAEs, we proposed a new data selection algorithm for instruction tuning corset selection. The models fine-tuned on our selected data consistently outperform other selection methods in instruction-following abilities across different models and datasets. Also, when we change the ratio of selected data, our approach consistently achieves better results. Besides, we can use our data diversity measure to explain why longer instruction-response data usually leads to better model performance. In the future, we hope to extend this pipeline to other directions such as preference data selection, or how to make model safer through data selection.
 
\section*{Impact Statement}
Our research has the potential to impact the fields of finetuning LLMs by introducing a novel measure of data diversity through learned monosemanticity in sparse autoencoders. The proposed data selection algorithm for instruction tuning corset selection can lead to improved model performance, increased efficiency, and enhanced reliability.

\textbf{Improved Model Performance}: Our approach consistently outperforms other selection methods in instruction-following abilities across different models and datasets, resulting in better overall model performance.

\textbf{Increased Efficiency}: By selecting the most relevant data, our method reduces the amount of data required for fine-tuning, leading to faster training times and lower computational costs.

\textbf{Enhanced Reliability}: Our data diversity measure provides insights into why longer instruction-response data typically leads to better model performance, enabling more informed decisions about data selection and model development.

\textbf{Broader Applications}: Our pipeline has the potential to be extended to other areas, such as preference data selection and enhancing model safety through data selection, further expanding its impact on the field.

\clearpage
\newpage
\bibliographystyle{assets/plainnat}
\bibliography{example_paper}

\clearpage
\newpage
\beginappendix

\section{Models and Datasets Links}\label{app:links}
In Table \ref{tab:models-links}, we summarize all the open-source links for the models, datasets and codebases used in our work.
\begin{table}[h]
    \centering
    \begin{tabular}{l|l}
        \hline
        Models, Datasets, Codebase & Links \\ \hline
        Llama-2-13b-base &  https://huggingface.co/meta-llama/Llama-2-13b \\ \hline
        Gemma-2-9b &  https://huggingface.co/google/gemma-2-9b \\ \hline
        Llama-2-7b-base &  https://huggingface.co/meta-llama/Llama-2-7b \\ \hline
        Alpaca &  https://huggingface.co/datasets/tatsu-lab/alpaca \\ \hline
        stanford\_alpaca &  https://github.com/tatsu-lab/stanford\_alpaca \\ \hline
        Llama-3.1-8B-Instruct &  https://huggingface.co/meta-llama/Llama-3.1-8B-Instruct \\ \hline
        lm-evaluation-harness &  https://github.com/EleutherAI/lm-evaluation-harness \\ \hline
        Gemma-SAE &  https://huggingface.co/google/gemma-scope-9b-pt-res \\ \hline
        WizardLM\_evol\_instruct\_70k &  
        https://huggingface.co/datasets/WizardLMTeam/WizardLM\_evol\_instruct\_70k \\ \hline
    \end{tabular}
    \caption{Models and Datasets Links}
    \label{tab:models-links}
\end{table}

\section{Benchmark}
Here we show the complete results of all model performance on the benchmarks.

\begin{table*}[htbp]
    \centering
    \begin{tabular}{c|c|ccccc|ccccc}
        \toprule
        \multirow{2}{*}{Data} &\multirow{2}{*}{Method} & \multicolumn{5}{c|}{\textbf{Alpaca-52k}} & \multicolumn{5}{c}{\textbf{WizardLM\_evol\_instruct-70k}} \\ \cline{3-12}
        & & \tiny{MMLU} & \tiny{Winogrande} & \tiny{Truthfulqa} & \tiny{ARC} & \tiny{Gsm8k} & \tiny{MMLU} & \tiny{Winogrande} & \tiny{Truthfulqa} & \tiny{ARC} & \tiny{Gsm8k} \\ \cline{1-12}
        All & NA & 53.33 & 71.98 & 34.88 & 51.37 & 13.87  & 55.01  & 74.59 & 39.05 & 53.58 & 27.90 \\ \cline{1-12}
        \multirow{6}{*}{1k} 
        &\#INSTAG & 53.59 & 70.64 & 33.41 & 48.81 & 15.09 & 53.72 & 71.82 & 35.99 & 50.77 & 21.76 \\ 
        &Longest-instruction & 53.29 & 70.13 & 33.61 & 48.92 & 21.23 & 54.97 & 73.09 & 37.58 & 50.77 & 24.03 \\  
        & Longest-response  & 53.16 & 74.59 & 35.37 & 52.22 & 21.38 & 54.81 & 74.82 & 38.07 & 52.47 & 26.61 \\ 
        & Repr Filter &  53.81 & 73.48 & 35.13  & 50.77 & 11.75 & 50.95 & 74.03 & 36.60 & 47.70 & 24.26 \\  \cline{2-12}
        &\textbf{\one} & 54.49 & 74.19 & 33.17 & 51.71 & 18.27 & 54.88 & 74.59 & 38.43 & 55.80 & 25.25 \\ \cline{2-12}
        & \textbf{\two}  & 54.09 & 73.40 & 33.28 & 52.82 & 21.30 & 54.47 & 73.24 & 37.94 & 51.02 & 26.23 \\ 
        \midrule
        \multirow{6}{*}{3k}
        &\#INSTAG & 52.78 & 69.38 & 33.66 & 48.21 & 13.57 & 55.19 & 73.32 & 39.17 & 48.89 & 22.44 \\ 
        &Longest-instruction & 55.21 & 72.93 & 33.29 & 50.68 & 16.60 & 55.31 & 73.72 & 40.15 & 51.71 & 23.88 \\  
        & Longest-response  & 53.26 & 70.80 & 32.56 & 49.49 & 20.70 & 54.05 & 74.74 & 37.94 & 51.11 & 21.91 \\          
        & Repr Filter & 54.11 & 70.72 & 33.66  & 49.06 & 13.57 & 55.21 & 75.06 & 38.68 & 50.60 & 21.68 \\  \cline{2-12}
        &\textbf{\one} & 54.17 & 71.43 & 33.05 & 78.83 & 15.16 & 55.05 & 74.43 & 37.45 & 53.84 & 22.35 \\ \cline{2-12}
        & \textbf{\two}  & 54.39 & 70.56 & 32.93 & 52.47 & 11.90 & 55.38 & 74.19 & 36.35 & 53.41 & 24.41 \\ 
        \midrule
        \multirow{6}{*}{5k}
        &\#INSTAG & 50.19 & 69.23 & 29.87 & 47.01 & 13.04 & 55.13 & 74.03 & 38.80 & 78.24 & 22.74 \\ 
        &Longest-instruction & 52.39 & 70.96 & 29.50 & 50.43 & 13.34 & 55.04 & 74.27 & 38.31 & 81.86 & 23.81 \\  
        & Longest-response  & 51.09 & 71.02 & 32.24 & 49.87 & 12.59 & 54.73 & 74.66 & 37.33 & 53.24 & 20.47 \\  
        & Repr Filter & 54.51 & 73.95 & 34.52  & 54.01 & 16.60 & 55.80 & 75.53 & 37.70 & 52.99  & 25.25 \\  \cline{2-12}
        &\textbf{\one} & 51.65 & 70.32 & 32.44 & 48.63 & 11.37 & 55.23 & 73.95 & 37.70 & 51.79 & 23.20 \\ \cline{2-12}
        & \textbf{\two}  & 52.22 & 71.59 & 31.21 & 49.66 & 11.60 & 54.52 & 73.64 & 38.19 & 53.16 & 23.81 \\ 
        \bottomrule
    \end{tabular}
    \caption{Comparison of different models fine-tuned from Llama-2-13b using 1k, 3k and 5k data selected from Alpaca and WizardLM\_evol\_instruct\_70k datasets. '-' represents not appliable.}\label{tab: benchmark}
\end{table*}

\section{Additional Case Study}\label{app: case_study}
\begin{figure*}[h]
\centering
\includegraphics[width=0.95\linewidth]{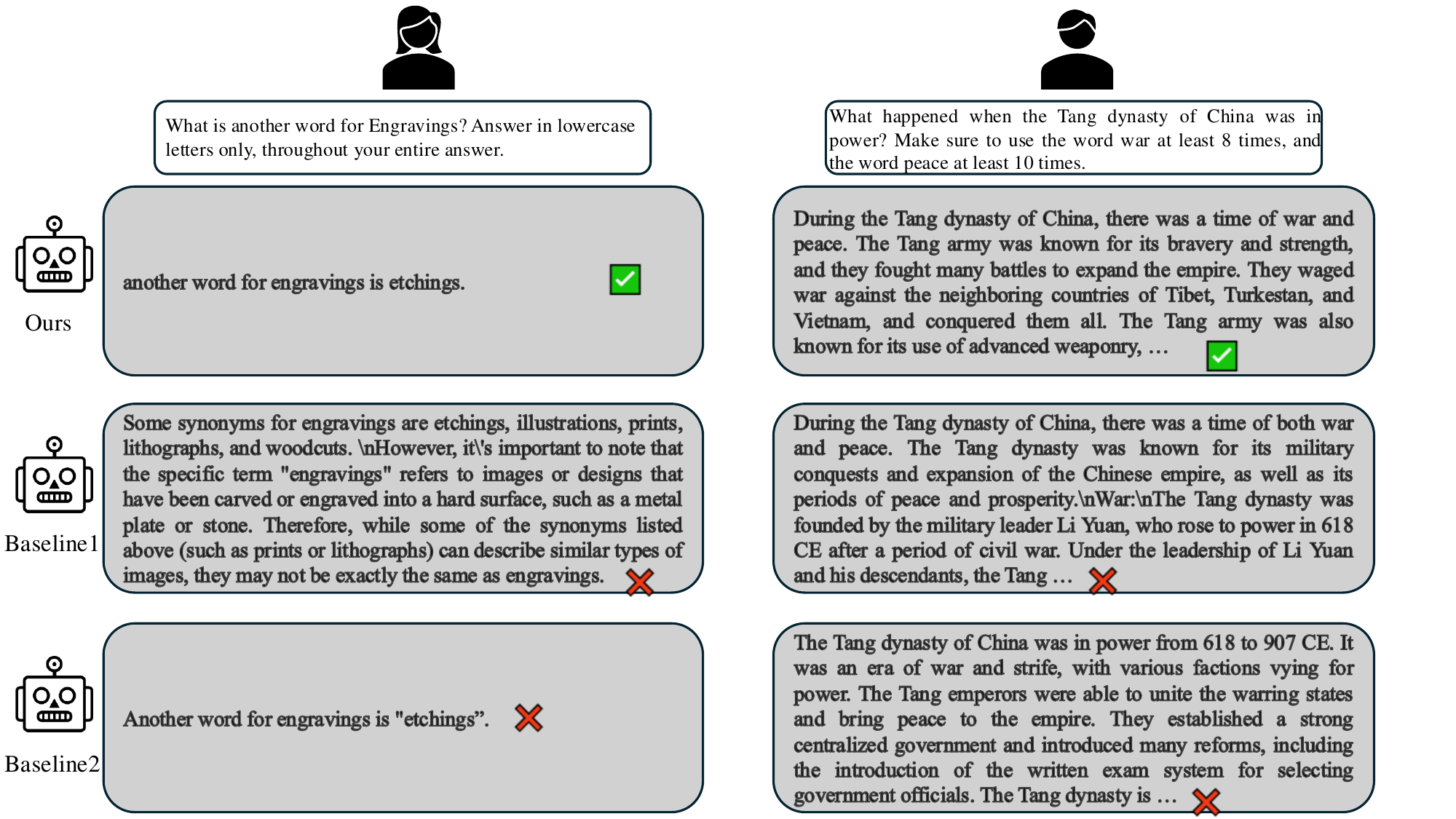}
\caption{Case study to illustrate the
instruction-following performance of Llama-2-13B model finetuned
on $3$k data selected from Alpaca with our \one method, \textit{\#INSTAG} (Baseline1) and Longest-response (Baseline2).}\label{fig: case_study2}
\end{figure*}

Here we show additional examples in Figure \ref{fig: case_study2} as case studies, and the instructions are picked from the IFEval dataset. Our model outperforms both baselines in instruction adherence and generation correctness.

\begin{figure*}[h]
\centering
\includegraphics[width=0.95\linewidth]{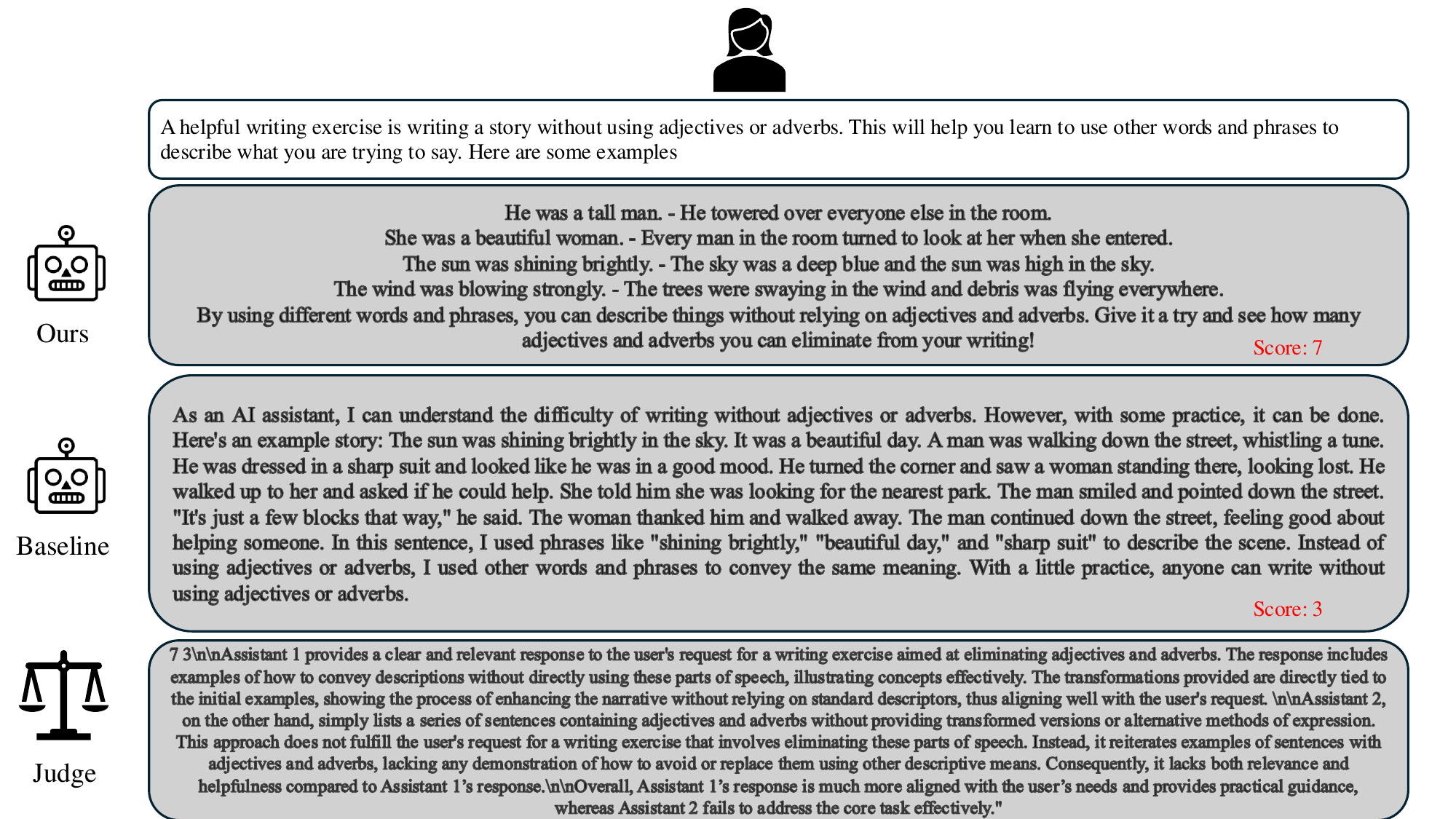}
\caption{Case study to illustrate the instruction-following performance of Llama-2-13B model fine-tuned on $5$k data selected from WizardLM\_evol\_instruct-70k with our \one method, and \textit{Repr Filter} (Baseline), with GPT4o-as-the-Judge.}\label{fig: case_study3}
\end{figure*}

\begin{figure*}[h]
\centering
\includegraphics[width=0.95\linewidth]{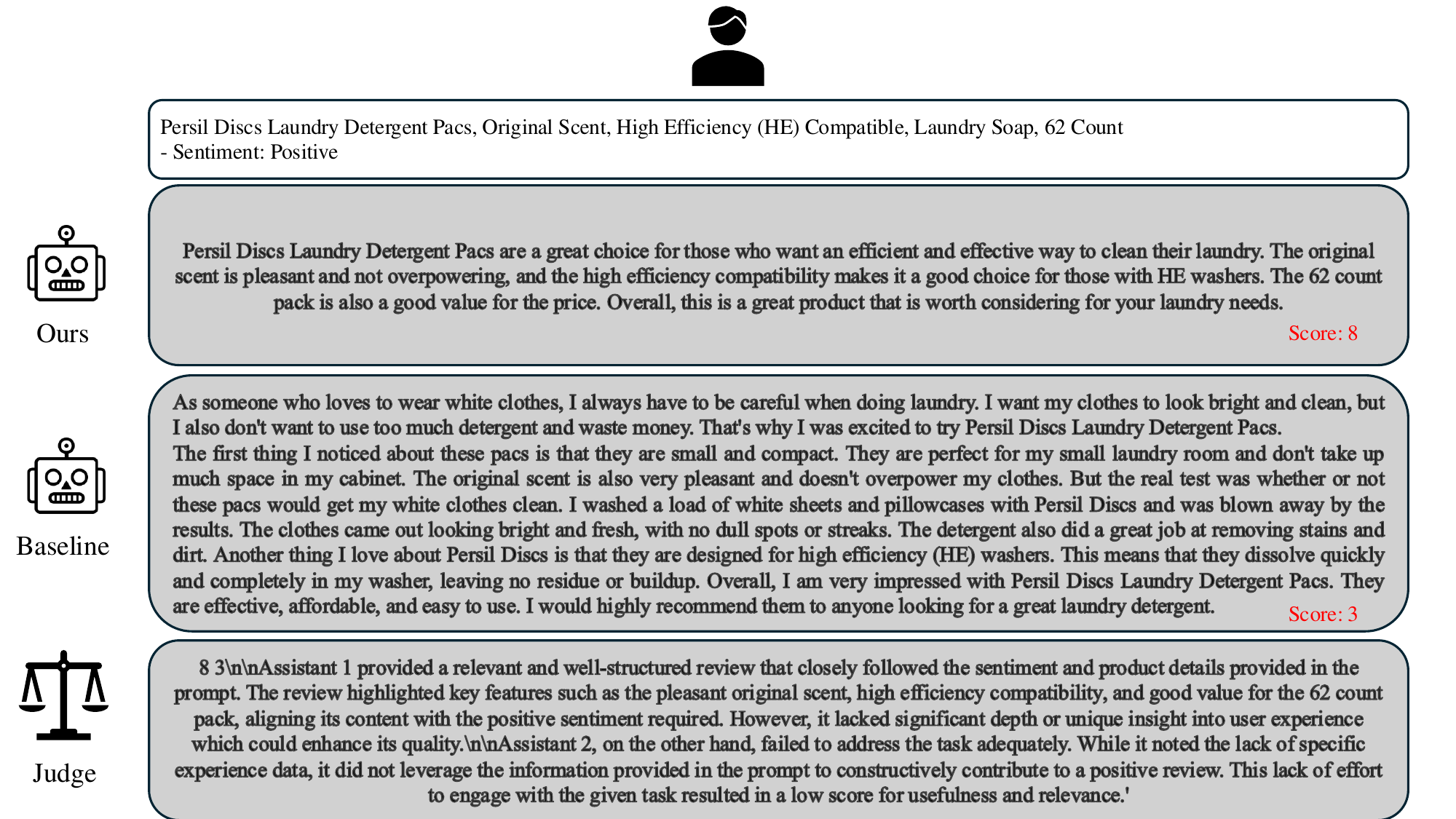}
\caption{Case study to illustrate the instruction-following performance of Llama-2-13B model fine-tuned on $5$k data selected from WizardLM\_evol\_instruct-70k with our \one method, and \textit{Repr Filter} (Baseline), with GPT4o-as-the-Judge.}\label{fig: case_study4}
\end{figure*}

Besides, we also show additional results in Figures \ref{fig: case_study3} and \ref{fig: case_study4} with instructions from the AlpacaEval 2.0 benchmark to show the general instruction following abilities. 

\section{Additional SAE Training Loss VS. Batch Size}\label{app: sae-bs}
\begin{figure}[h]
\centering
\includegraphics[width=0.85\linewidth]{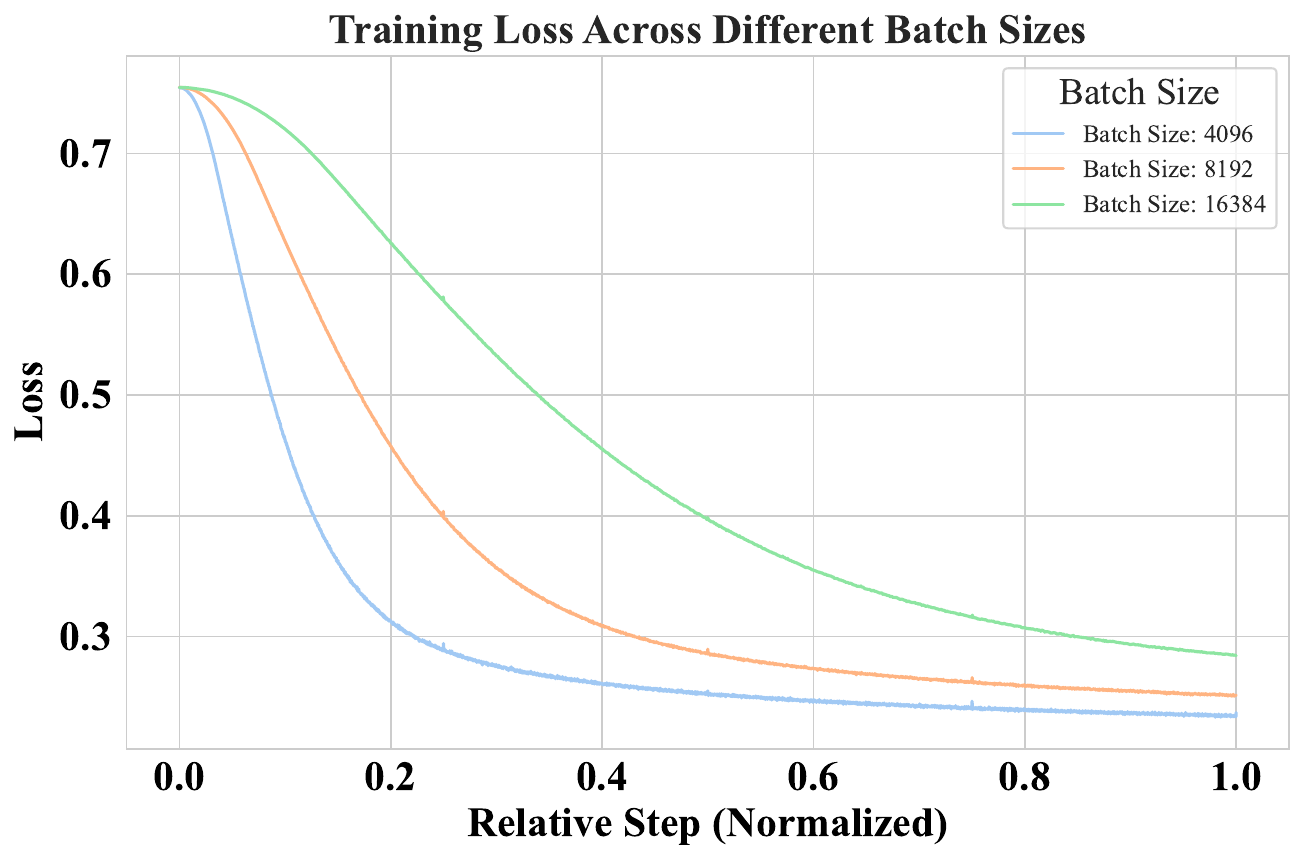}
\caption{The training loss of TopK-SAE on the layer 31 of Llama-3.1-8b-instruct.}\label{fig: loss_bs}
\end{figure}

Here in figure \ref{fig: loss_bs} we the SAE training loss for layer 29 when we change the batch size from $4,096$, $8,192$ to $16,384$. We can see that setting batch size to $4,096$ makes the SAE training convergence faster and leads to smaller final loss.

\section{Instruction Following Evaluation}\label{app: comparison}
\begin{figure}[h]
\centering
\includegraphics[width=0.9\linewidth]{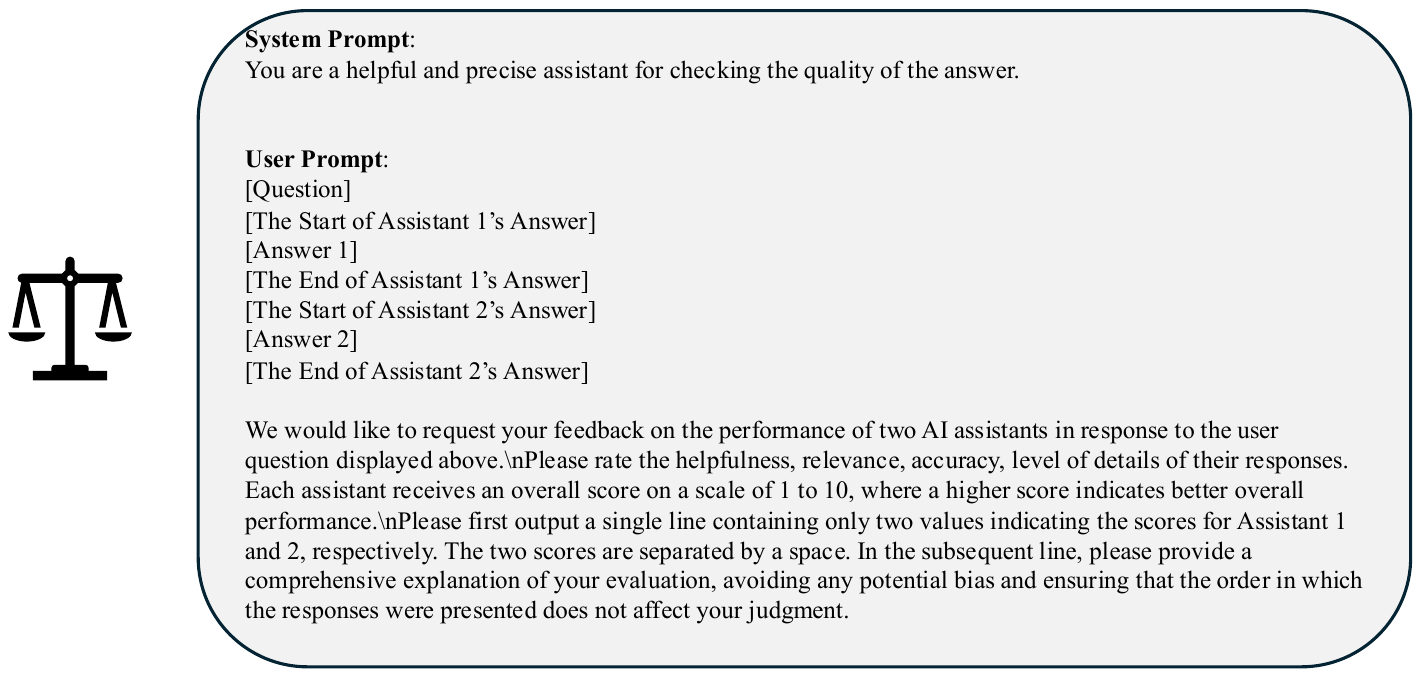}
\caption{The prompt template used for GPT-4o as the judge.}\label{fig: evaluation_prompt}
\end{figure}

The head-to-head comparison was conducted on the 805 instructions in the AlpacaEval dataset.
We utilize the same evaluation prompt template for GPT-4o, as employed by AlpaGasus \citep{chenalpagasus}, Longest \citep{zhaolong} and originally used in Vicuna work \citep{chiang2023vicuna}
. In Figure \ref{fig: evaluation_prompt}, we show the evaluation prompt for GPT-4o. In our preliminary experiments, we find that the the response length has a significant correlation with model judge, aligning with the findings in \citep{dubois2024length}. Thus, we first truncate both responses to the same length and then continue the LLM-as-a-judge for head-to-head comparison. For human evaluators, we hire PhD volunteers to only evaulate 200 of the 805 instructions and only ask for a score, with no explanation for simplicity. On the other hand, for the AlpacaEval 2.0 leadboard evaluation, we keep the original response length since they also offer the length controlled win rate.

\end{document}